\documentclass[11pt,a4paper]{article}
\usepackage{naaclhlt2018}
\usepackage{times}
\usepackage{latexsym}
\usepackage{booktabs}
\usepackage{todonotes}

\usepackage{url}

\usepackage[usenames,dvipsnames]{pstricks}
\usepackage{epsfig}

\usepackage{enumitem} 
\usepackage[hang,flushmargin]{footmisc} 

\aclfinalcopy 
\def\aclpaperid{3039} 

\title{A Deeper Look into Dependency-Based Word Embeddings}

\author{Sean MacAvaney \\
  Information Retrieval Lab \\
  Department of Computer Science \\
  Georgetown University \\
  Washington, DC, USA\\
  {\tt sean@ir.cs.georgetown.edu} \\\And
  Amir Zeldes \\
  Department of Linguistics \\Georgetown University \\
  Washington, DC, USA\\
  {\tt amir.zeldes@georgetown.edu} \\
  }

\date{}

\begin{document}
\maketitle
\begin{abstract}
We investigate the effect of various dependency-based word embeddings on distinguishing between functional and domain similarity, word similarity rankings, and two downstream tasks in English. Variations include word embeddings trained using context windows from Stanford and Universal dependencies at several levels of enhancement (ranging from unlabeled, to Enhanced++ dependencies). Results are compared to basic linear contexts and evaluated on several datasets. We found that embeddings trained with Universal and Stanford dependency contexts excel at different tasks, and that enhanced dependencies often improve performance.
\end{abstract}

\section{Introduction}

For many natural language processing applications, it is
important to understand word-level semantics. Recently, word embeddings trained with neural networks have gained popularity~\cite{mikolov2013efficient,pennington2014glove}, and have been successfully used for various tasks, such as machine translation~\cite{zou2013bilingual} and information retrieval~\cite{Hui2017PACRRAP}.

Word embeddings are usually trained using linear bag-of-words contexts, i.e. tokens positioned around a word are used to learn a dense representation of that word. \citet{levy2014dependency} challenged the use of linear contexts, proposing instead to use contexts based on dependency parses. (This is akin to prior work that found that dependency contexts are useful for vector models~\cite{pado07:_depen,Baroni:2010:DMG:1945043.1945049}.) They found that embeddings trained this way are better at capturing semantic similarity, rather than relatedness. For instance, embeddings trained using linear contexts place \textit{Hogwarts} (the fictional setting of the Harry Potter series) near \textit{Dumbledore} (a character from the series), whereas embeddings trained with dependency contexts place \textit{Hogwarts} near \textit{Sunnydale} (fictional setting of the series Buffy the Vampire Slayer). The former is \textit{relatedness}, whereas the latter is \textit{similarity}.

Work since \citet{levy2014dependency} examined the use of dependency contexts and sentence feature representations for sentence classification~\cite{komninos2016dependency}. \citet{Li2017InvestigatingDC} filled in research gaps relating to model type (e.g., CBOW, Skip-Gram, GloVe) and dependency labeling. Interestingly, \citet{abnar2018experiential} recently found that dependency-based word embeddings excel at predicting brain activation patterns. The best model to date for distinguishing between similarity and relatedness combines word embeddings, WordNet, and dictionaries~\cite{Recski:2016c}.

One limitation of existing work is that it has only explored one dependency scheme: the English-tailored Stanford Dependencies~\cite{de2008stanford}. We provide further analysis using the cross-lingual Universal Dependencies~\cite{nivre2016universal}. Although we do not compare cross-lingual embeddings in our study, we will address one important question for English: are Universal Dependencies, which are less tailored to English, actually better or worse than the English-specific labels and graphs?
Furthermore, we investigate approaches to simplifying and extending dependencies, including Enhanced dependencies and Enhanced++ dependencies ~\cite{schuster2016enhanced}, as well as two levels of relation simplification. We hypothesize that the cross-lingual generalizations from universal dependencies and the additional context from enhanced dependencies should improve the performance of word embeddings at distinguishing between functional and domain similarity. We also investigate how these differences impact word embedding performance at word similarity rankings and two downstream tasks: question-type classification and named entity recognition.

\section{Method}

\begin{figure}
\centering
%
%

\psscalebox{1.0 1.0} 
{
\begin{pspicture}(0,-3.2775)(7.0,3.2775)
\psframe[linecolor=black, linewidth=0.04, dimen=outer](6.6,2.4775)(3.8,-1.3225)
\rput(5.2,2.8775){Enhanced + +}
\psframe[linecolor=black, linewidth=0.04, dimen=outer](6.4,1.6775)(4.0,-1.1225)
\psframe[linecolor=black, linewidth=0.04, dimen=outer](6.2,0.8775)(4.2,-0.9225)
\psframe[linecolor=black, linewidth=0.04, dimen=outer](6.0,0.0775)(4.4,-0.7225)
\rput(5.2,2.0775){Enhanced}
\rput(5.2,1.2775){Basic}
\rput(5.2,0.4775){Simplified}
\psframe[linecolor=black, linewidth=0.04, dimen=outer](5.0,-1.9225)(3.4,-2.7225)
\psframe[linecolor=black, linewidth=0.04, dimen=outer](7.0,-1.9225)(5.4,-2.7225)
\rput(4.2,-2.3225){Stanford}
\rput(6.2,-2.3225){Universal}
\rput[b](5.2,-1.9225){+}
\psframe[linecolor=black, linewidth=0.04, dimen=outer](2.6,-0.7225)(0.2,-1.5225)
\psframe[linecolor=black, linewidth=0.04, dimen=outer](1.2,-1.9225)(0.0,-2.7225)
\psframe[linecolor=black, linewidth=0.04, dimen=outer](2.8,-1.9225)(1.6,-2.7225)
\rput[b](1.4,-1.9225){+}
\rput(1.4,-1.1225){CBOW}
\rput(0.6,-2.3225){k = 2}
\rput(2.2,-2.3225){k = 5}
\rput(1.4,-3.1225){Linear Contexts}
\rput(5.2,-3.1225){Dependency Contexts}
\psframe[linecolor=black, linewidth=0.04, dimen=outer](6.8,3.2775)(3.6,-1.5225)
\rput(5.2,-0.3225){\psscalebox{0.9}{Unlabeled}}
\psframe[linecolor=black, linewidth=0.04, dimen=outer](2.6,2.0775)(0.2,1.2775)
\rput(1.4,1.6775){Skip-Gram}
\psframe[linecolor=black, linewidth=0.04, dimen=outer](2.2,0.8775)(0.6,0.0775)
\rput(1.4,0.4775){Subword}
\rput(1.4,1.0775){+}

\pscustom[linecolor=black, linewidth=0.04]
{
\newpath
\moveto(2.8,-1.3225)
}
\psline[linecolor=black, linewidth=0.04, linestyle=dotted, dotsep=0.10583334cm](0.3,-0.3225)(2.6,-0.3225)
\psline[linecolor=black, linewidth=0.04, linestyle=dotted, dotsep=0.10583334cm](3.1,-3.2225)(3.1,3.0775)
\end{pspicture}
}
\caption{Visual relationship between types of embedding contexts. Each layer of enhancement adds more information to the dependency context (e.g., simplified adds dependency labels to the unlabeled context). We investigate CBOW using both a context window of $k=2$ and $k=5$, and we use the SkipGram model both with and without subword information.}
\label{fig_rel}
\end{figure}
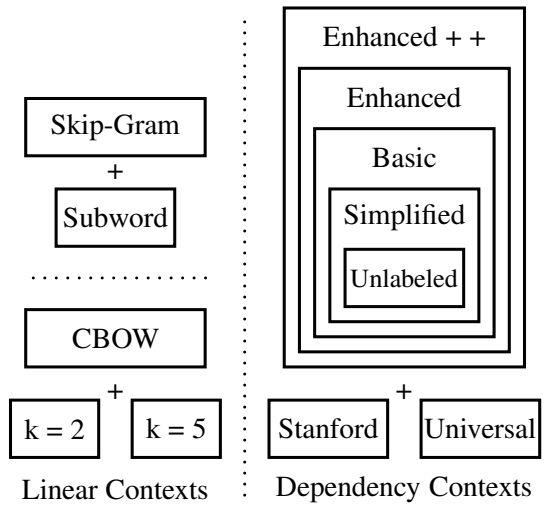

In this work, we explore the effect of two dependency annotation schemes on the resulting embeddings. Each scheme is evaluated in five levels of enhancement. These embeddings are compared to embeddings trained with linear contexts using the continuous bag of words (CBOW) with a context window of $k=2$ and $k=5$, and Skip-Gram contexts with and without subword information. These configurations are summarized in Figure~\ref{fig_rel}.

Two dependency annotation schemes for English are Stanford dependencies~\cite{de2008stanford} and Universal dependencies~\cite{nivre2016universal}. Stanford dependencies are tailored to English text, including dependencies that are not necessarily relevant cross-lingually (e.g. a label \textit{prt} for particles like \textit{up} in \textit{pick up}). Universal dependencies are more generalized and designed to work cross-lingually. Many structures are similar between the two schemes, but important differences exist. For instance, in Stanford dependencies, prepositions head their phrase and depend on the modified word (\textit{in} is the head of \textit{in Kansas}), whereas in universal dependencies, prepositions depend on the prepositional object (\textit{Kansas} dominates \textit{in}). Intuitively, these differences should have a moderate effect on the resulting embeddings because different words will be in a given word's context.

\begin{table}
\centering\small
\begin{tabular}{p{0.7cm}p{6cm}}
\toprule
Simp. & Basic                                                                                                                                              \\ \midrule
\multicolumn{2}{l}{\bf Stanford dependencies}                                                                                                                                     \\
mod              & poss, prt, predet, det, amod, tmod, npadvmod, possessive, advmod, quantmod, preconj, mark, vmod, nn, num, prep, appos, mwe, mod, number, neg, advcl, rcmod \\
arg              & agent, iobj, dobj, acomp, pcomp, pobj, ccomp, arg, subj, csubj, obj, xcomp, nsubj                                              \\
aux              & aux, cop                                                                                                                                          \\
sdep             & xsubj, sdep                                                                                                                                                \\ \midrule
\multicolumn{2}{l}{\bf Universal dependencies}                                                                                                                                    \\
core             & iobj, dobj, ccomp, csubj, obj, xcomp, nsubj                                                                                          \\
ncore            & discourse, cop, advmod, dislocated, vocative, aux, advcl, mark, obl, expl                                                                         \\
nom              & case, nmod, acl, neg, appos, det, amod, nummod                                                                                                        \\
coord            & cc, conj                                                                                                                                                   \\
special          & goeswith, reparandum, orphan                                                                                                                               \\
loose            & parataxis, list                                                                                                                                            \\
mwe              & compound, mwe, flat                                                                                                                                 \\
other            & punct, dep, root \\
\bottomrule
\end{tabular}
\caption{Simplified Stanford and Universal dependency labels. For simplified dependencies, basic labels are collapsed into the simplified label shown in this table. (Relations not found in this table were left as is.)}
\label{tab_simp}
\end{table}


We also investigate five levels of enhancement for each dependency scheme. \textit{Basic} dependencies are the core dependency structure provided by the scheme. \textit{Simplified} dependencies are more coarse basic dependencies, collapsing similar labels into rough classes. The categories are based off of the Stanford Typed Dependencies Manual~\cite{de2008stanford2} and the Universal Dependency Typology~\cite{de2014universal}, and are listed in Table~\ref{tab_simp}. Note that the two dependency schemes organize the relations in different ways, and thus the two types of simplified dependencies represent slightly different structures. The \textit{unlabeled} dependency context removes all labels, and just captures syntactically adjacent tokens.

\textit{Enhanced} and \textit{Enhanced++} dependencies~\cite{schuster2016enhanced} address some practical dependency distance issues by extending basic dependency edges. Enhanced dependencies augment modifiers and conjuncts with their parents' labels, propagate governors and dependents for indirectly governed arguments, and add subjects to controlled verbs. Enhanced++ dependencies allow for the deletion of edges to better capture English phenomena, including partitives and light noun constructions, multi-word prepositions, conjoined prepositions, and relative pronouns.

\section{Experimental Setup}

We use the Stanford CoreNLP parser\footnote{\url{stanfordnlp.github.io/CoreNLP/}} to parse basic, Enhanced, and Enhanced++ dependencies. We use the Stanford english\_SD model to parse Stanford dependencies (trained on the Penn Treebank) and english\_UD model to parse  Universal dependencies (trained on the Universal Dependencies Corpus for English). We acknowledge that differences in both the size of the training data (Penn Treebank is larger than the Universal Dependency Corpus for English), and the accuracy of the parse can have an effect on our overall performance. We used our own converter to generate simple dependencies based on the rules shown in Table~\ref{tab_simp}. We use the modified \texttt{word2vecf} software\footnote{\url{bitbucket.org/yoavgo/word2vecf}}~\citet{levy2014dependency} that works with arbitrary embedding contexts to train dependency-based word embeddings.

As baselines, we train the following linear-context embeddings using the original \texttt{word2vec} software:\footnote{\url{code.google.com/p/word2vec/}} CBOW with \mbox{$k=2$}, CBOW with \mbox{$k=5$}, and Skip-Gram. We also train enriched Skip-Gram embeddings including subword information~\cite{bojanowski2016enriching} using \mbox{fastText}.\footnote{\url{github.com/facebookresearch/fastText}}

For all embeddings, we use a cleaned recent dump of English Wikipedia (November 2017, 4.3B tokens) as training data. We evaluate each on the following tasks:

\paragraph{Similarity over Relatedness} Akin to the quantitative analysis done by \citet{levy2014dependency}, we test to see how well each approach ranks similar items above related items. Given pairs of similar and related words, we rank each word pair by the cosine similarity of the corresponding word embeddings, and report the area-under-curve (AUC) of the resulting precision-recall curve. We use the labeled WordSim-353~\cite{agirre2009study,finkelstein2001placing} and the Chiarello dataset~\cite{chiarello1990semantic} as a source of similar and related word pairs. For WordSim-353, we only consider pairs with similarity/relatedness scores of at least 5/10, yielding 90 similar pairs and 147 related pairs. For Chiarello, we disregard pairs that are marked as both similar and related, yielding 48 similar pairs and 48 related pairs.

\paragraph{Ranked Similarity} This evaluation uses a list of word pairs that are ranked by degree of functional similarity. For each word pair, we calculate the cosine similarity, and compare the ranking to that of the human-annotated list using the Spearman correlation. We use SimLex-999~\cite{hill2016simlex} as a ranking of functional similarity. Since this dataset distinguishes between nouns, adjectives, and verbs, we report individual correlations in addition to the overall correlation.

\paragraph{Question-type Classification (QC)} We use an existing QC implementation\footnote{\url{github.com/zhegan27/sentence_classification}} that uses a bidirectional LSTM. We train the model with 20 epochs, and report the average accuracy over 10 runs for each set of embeddings. We train and evaluate using the TREC QC dataset~\cite{li2002learning}. We modified the approach to use fixed (non-trainable) embeddings, allowing us to compare the impact of each embedding type.

\paragraph{Named Entity Recognition (NER)} We use the \citet{2017neuroner} NER implementation\footnote{\url{github.com/Franck-Dernoncourt/NeuroNER}} that uses a bidirectional LSTM. Training consists of a maximum of 100 epochs, with early stopping after 10 consecutive epochs with no improvement to validation performance. We evaluate NER using the F1 score on the CoNLL NER dataset~\cite{tjong2003introduction}. Like the QC task, we use a non-trainable embedding layer.

\section{Results}

\subsection{Similarity over Relatedness}

\setlength{\tabcolsep}{5.5pt}
\begin{table*}
\centering
\begin{tabular}{lrrrrrrrrrr}
\toprule
& \multicolumn{2}{c}{(a) Sim/rel (AUC)} & & \multicolumn{4}{c}{(b) Ranked sim (Spearman)} & & \multicolumn{2}{c}{(c) Downstream} \\ \cmidrule{2-3} \cmidrule{5-8} \cmidrule{10-11}
Embeddings & WS353 & Chiarello & & Overall & Noun & Adj. & Verb & & QC (Acc) & NER (F1) \\
\midrule\multicolumn{6}{l}{\bf Universal embeddings} \\
Unlabeled & 0.786 & 0.711 & & 0.370 & 0.408 & 0.484 & 0.252 &  & 0.915 & 0.877 \\ 
Simplified & 0.805 & 0.774 & & 0.394 & 0.420 & 0.475 & 0.309 &  & 0.913 & 0.870 \\ 
Basic & 0.801 & 0.761 & & 0.391 & 0.421 & 0.451 & 0.331 &  & 0.920 & 0.876 \\ 
Enhanced & \bf0.823 & \bf0.792 & & 0.398 & 0.416 & 0.473 & \bf0.350 &  & 0.915 & 0.875 \\ 
Enhanced++ & 0.820 & 0.791 & & 0.396 & 0.416 & 0.461 & 0.348 &  & 0.917 & 0.882 \\ 
\midrule\multicolumn{6}{l}{\bf Stanford embeddings} \\
Unlabeled & 0.790 & 0.741 & & 0.382 & 0.414 & \bf0.507 & 0.256 &  & 0.911 & 0.870 \\ 
Simplified & 0.793 & 0.748 & & 0.393 & 0.416 & 0.501 & 0.297 &  &\bf0.923 & 0.873 \\ 
Basic & 0.808 & 0.769 & &\bf0.402 & \bf0.422 & 0.494 & 0.341 & & 0.910 & 0.865 \\ 
Enhanced & 0.817 & 0.755 & & 0.399 & 0.420 & 0.482 & 0.338 &  & 0.911 & 0.871 \\ 
Enhanced++ & 0.810 & 0.764 & & 0.398 & 0.417 & 0.496 & 0.346 &  & 0.918 & 0.878 \\ 
\midrule\multicolumn{6}{l}{\bf Baselines (linear contexts)} \\
CBOW, k=2 & 0.696 & 0.537 & & 0.311 & 0.355 & 0.338 & 0.252 &  & 0.913 & 0.885 \\ 
CBOW, k=5 & 0.701 & 0.524 & & 0.309 & 0.353 & 0.358 & 0.258 &  & 0.899 & 0.893 \\ 
Skip-Gram & 0.617 & 0.543 & & 0.264 & 0.304 & 0.368 & 0.135 &  & 0.898 & 0.881 \\
SG + Subword & 0.615 & 0.456 & & 0.324 & 0.358 & 0.451 & 0.166 &  & 0.897 &\bf0.887 \\ 
\bottomrule
\end{tabular}
\caption{Results of various dependency-based word embeddings, and baseline linear contexts at (a) similarity over relatedness, (b) ranked similarity, and (c) downstream tasks of question classification and named entity recognition.}
\label{tab_results}
\end{table*}

The results for the WordSim-353 (WS353) and Chiarello datasets are given in Table~\ref{tab_results}a.  For the WS353 evaluation, notice that the Enhanced dependencies for both Universal and Stanford dependencies outperform the others in each scheme. Even the poorest-performing level of enhancement (unlabeled), however, yields a considerable gain over the linear contexts. Both Skip-Gram variants yield the worst performance, indicating that they capture relatedness better than similarity. For the Chiarello evaluation, the linear contexts perform even worse, while the Enhanced Universal embeddings again outperform the other approaches.

These results reinforce the~\citet{levy2014dependency} findings that dependency-based word embeddings do a better job at distinguishing similarity rather than relatedness because it holds for multiple dependency schemes and levels of enhancement. The Enhanced universal embeddings outperformed the other settings for both datasets. For Chiarello, the margin between the two is statistically significant, whereas for WS353 it is not. This might be due to the fact that the the Chiarello dataset consists of manually-selected pairs that exhibit similarity or relatedness, whereas the settings for WS353 allow for some marginally related or similar terms through (e.g., \textit{size} is related to \textit{prominence}, and \textit{monk} is similar to \textit{oracle}).


\subsection{Ranked Similarity}

Spearman correlation results for ranked similarity on the SimLex-999 dataset are reported in Table~\ref{tab_results}b. \textit{Overall} results indicate the performance on the entire collection. In this environment, basic Stanford embeddings outperform all other embeddings explored. This is an interesting result because it shows that the additional dependency labels added for Enhanced embeddings (e.g. for conjunction) do not improve the ranking performance. This trend does not hold for Universal embeddings, with the enhanced versions outperforming the basic embeddings.

All dependency-based word embeddings significantly outperform the baseline methods (10 folds, paired t-test, $p<0.05$). Furthermore, the unlabeled Universal embeddings performed significantly worse than the simplified Universal, and the simplified, basic, and Enhanced Stanford dependencies, indicating that dependency labels are important for ranking.

Table~\ref{tab_results}b also includes results for word pairs by part of speech individually. As the majority category, Noun-Noun scores ($n=666$) mimic the behavior of the overall scores, with basic Stanford embeddings outperforming other approaches. Interestingly, Adjective-Adjective pairs ($n=111$) performed best with unlabeled Stanford dependencies. Since unlabeled also performs best among universal embeddings, this indicates that dependency labels are not useful for adjective similarity, possibly because adjectives have comparatively few ambiguous functions. Verb-Verb pairs ($n=222$) performed best with Enhanced universal embeddings. This indicates that the augmentation of governors, dependents, and subjects of controlled verbs is particularly useful given the universal dependency scheme, and less so for the English-specific Stanford dependency scheme. Both Stanford and universal unlabeled dependencies performed significantly worse compared to all basic, Enhanced, and Enhanced++ dependencies (5 folds, paired t-test, $p<0.05$). This indicates that dependency labels are particularly important for verb similarity.

\subsection{Downstream Tasks}

\renewcommand{\arraystretch}{1.0}
\begin{table}
\centering
\small
\begin{tabular}{lrr}
\toprule
Embeddings & QC (Acc) & NER (F1) \\
\midrule\multicolumn{3}{l}{\bf Universal embeddings} \\
Unbound & 0.921 (+0.007) & 0.887 (+0.000) \\
Simplified & 0.929 (+0.016) & 0.883 (+0.013) \\
Basic & 0.920 (+0.000) & 0.891 (+0.015) \\
Enhanced & 0.923 (+0.008) & 0.886 (+0.010) \\
Enhanced++ & 0.927 (+0.010) & 0.890 (+0.008) \\
\midrule\multicolumn{3}{l}{\bf Stanford embeddings} \\
Unbound & 0.926 (+0.015) & 0.879 (+0.009) \\
Simplified &\bf0.933 (+0.010) & 0.877 (+0.004) \\
Basic & 0.927 (+0.017) & 0.885 (+0.020) \\
Enhanced & 0.923 (+0.013) & 0.885 (+0.014) \\
Enhanced++ & 0.929 (+0.011) & 0.884 (+0.006) \\
\midrule\multicolumn{3}{l}{\bf Baselines (linear contexts)} \\
CBOW, k=2 & 0.921 (+0.008) & 0.892 (+0.007) \\
CBOW, k=5 & 0.925 (+0.026) & 0.892 (+0.001) \\
Skip-Gram & 0.914 (+0.016) & 0.887 (+0.006) \\
SG + Subword & 0.919 (+0.022) &\bf0.896 (+0.009) \\
\bottomrule
\end{tabular}
\caption{Performance results when embeddings are further trained for the particular task. The number in parentheses gives the performance improvement compared to when embeddings are not trainable (Table~\ref{tab_results}c).}
\label{tab_downstream}
\end{table}

We present results for question-type classification and named entity recognition in Table~\ref{tab_results}c. Neither task appears to greatly benefit from embeddings that favor similarity over relatedness or that can rank based on functional similarity effectively without the enhanced sentence feature representations explored by \citet{komninos2016dependency}. We compare the results using to the performance of models with embedding training enabled in Table~\ref{tab_downstream}. As expected, this improves the results because the training captures task-specific information in the embeddings. Generally, the worst-performing embeddings gained the most (e.g., CBOW $k=5$ for QC, and basic Stanford for NER). However, the simplified Stanford embeddings and the embeddings with subword information still outperform the other approaches for QC and NER, respectively. This indicates that the initial state of the embeddings is still important to an extent, and cannot be learned fully for a given task.

\section{Conclusion}

In this work, we expanded previous work by \citet{levy2014dependency} by looking into variations of dependency-based word embeddings. We investigated two dependency schemes: Stanford and Universal embeddings. Each scheme was explored at various levels of enhancement, ranging from unlabeled contexts to Enhanced++ dependencies. All variations yielded significant improvements over linear contexts in most circumstances. For certain subtasks (e.g. Verb-Verb similarity), enhanced dependencies improved results more strongly, supporting current trends in the universal dependency community to promote enhanced representations. Given the disparate results across POS tags, future work could also evaluate ways of using a hybrid approach with different contexts for different parts of speech, or using concatenated embeddings.


\bibliography{naaclhlt2018}
\bibliographystyle{acl_natbib}

\end{document}